\documentclass[preprint,12pt]{elsarticle}
\usepackage{amssymb}
\usepackage{color}
\usepackage{amsmath}
\usepackage{multirow}
\usepackage{verbatim}
\usepackage{geometry}

\usepackage[pagebackref=true,breaklinks=true,colorlinks,bookmarks=false]{hyperref}

\journal{Neurocomputing}

\begin{document}

\begin{frontmatter}




\title{Dynamic Supervisor for Cross-dataset Object Detection}


\author[1]{Ze Chen}
\author[5]{Zhihang Fu}
\author[5]{Jianqiang Huang}
\author[5]{Mingyuan Tao}
\author[5]{Shengyu Li}

\author[1]{Rongxin Jiang}
\author[1]{Xiang Tian}
\author[1]{Yaowu Chen\corref{cor1}}
\ead{cyw@mail.bme.zju.edu.cn}
\author[5]{Xian-Sheng Hua}

\cortext[cor1]{Corresponding author}

\affiliation[1]{organization={Zhejiang University},%Department and Organization
            addressline={No.38 Zheda Road}, 
            city={Hangzhou},
            postcode={310027},
            state={Zhejiang},
            country={China}}

%\affiliation[2]{organization={Zhejiang University Embedded System Engineering Research Center, Ministry of Education of China},%Department and Organization
            %addressline={No.38 Zheda Road}, 
            %city={Hangzhou},
            %postcode={310027}, 
            %state={Zhejiang},
            %country={China}}

%\affiliation[3]{organization={Zhejiang University, the State Key Laboratory of Industrial Control Technology},%Department and Organization
            %addressline={No.38 Zheda Road}, 
            %city={Hangzhou},
            %postcode={310027},
            %state={Zhejiang},
            %country={China}}

%\affiliation[4]{organization={Zhejiang Provincial Key Laboratory for Network Multimedia Technologies},%Department and Organization
            %addressline={No.38 Zheda Road}, 
            %city={Hangzhou},
            %postcode={310027}, 
            %state={Zhejiang},
            %country={China}}

\affiliation[5]{organization={Alibaba Group},%Department and Organization
            %addressline={}, 
            city={Hangzhou},
            postcode={311121}, 
            state={Zhejiang},
            country={China}}

\date{}

\begin{abstract}
The application of cross-dataset training in object detection tasks is complicated because the inconsistency in the category range across datasets transforms fully supervised learning into semi-supervised learning.
To address this problem, recent studies focus on the generation of high-quality missing annotations.
In this study, we first point out that it is not enough to generate high-quality annotations using a single model, which only looks once for annotations.
Through detailed experimental analyses, we further conclude that hard-label training is conducive to generating high-recall annotations, while soft-label training tends to obtain high-precision annotations.
Inspired by the aspects mentioned above, we propose a dynamic supervisor framework that updates the annotations multiple times through multiple-updated submodels trained using hard and soft labels.
In the final generated annotations, both recall and precision improve significantly through the integration of hard-label training with soft-label training.
Extensive experiments conducted on various dataset combination settings support our analyses and demonstrate the superior performance of the proposed dynamic supervisor.
\end{abstract}




\end{frontmatter}


\section{Introduction}\label{introduction}

Fully supervised learning~\cite{jordan1992forward} has dominated the field of computer vision for decades.
One of the most distinctive features of fully supervised learning is ``data-driven learning" in which a certain annotated dataset~\cite{bossard2014food,fei2004learning,li2014deepreid} defines the capability boundary of the trained model (the categories that are identifiable and those that are not). 
As a result, if one intends to expand the capability boundary, the training data must be significantly augmented with a larger category set, and this process is time-consuming and labor-intensive.
Therefore, cross-dataset training~\cite{cao2010cross,peng2016unsupervised,perrett2017recurrent,Zhao_UniDet_ECCV20} has attracted the attention of academics seeking to avoid the unignorable costs of establishing new datasets.
In cross-dataset training, several existing datasets only need to be merged and trained to expand the capability boundary to the union of their category sets without any extra image labeling costs.

\begin{figure}[t]
\begin{center}
   \includegraphics[width=1.0\linewidth]{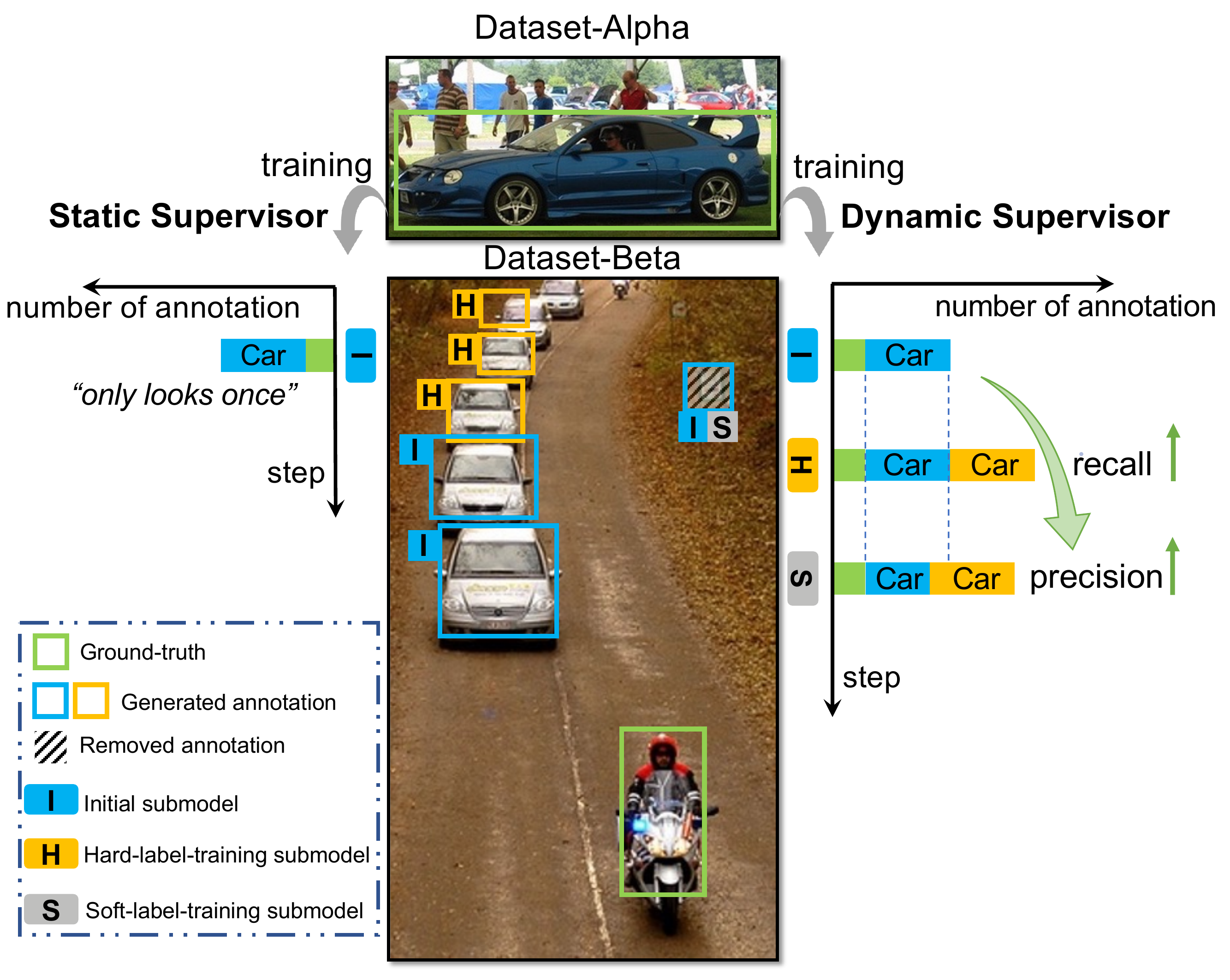}
\end{center}
   \caption{The difference between static supervisor and dynamic supervisor.
   There is no car category in the annotation of dataset Beta.
   Submodels trained on dataset Alpha are used to generate the missing annotations of dataset Beta.
   For the static supervisor, the initial submodel only looks once and detects three cars (one of them is false) as annotations.
   For the dynamic supervisor, after the annotation generation of the initial submodel, a hard-label-training submodel generates new annotations to increase the recall rate.
   Then, a soft-label-training submodel is utilized to filter out the unreliable annotations that increases the precision rate.
   }
\label{fig:fig1}
\end{figure}

Compared to classification~\cite{cao2010cross} and image retrieval~\cite{peng2016unsupervised} tasks, cross-dataset training on object detection is much more complicated.
This is because of the multi-label and multi-instance attributes of object detection.
In an image, detection annotation always contains multiple categories and instances.
A naive combination across detection datasets results in the incompleteness of the entire set.
For example, the category \textit{car}, which is annotated on dataset Alpha, is essentially not annotated at all on datasets Beta and Gamma.
These incomplete annotation sets further transform fully supervised learning into weakly~\cite{oquab2015object,tang2018pcl,jiang2021dynamic,yi2020wsodpb} or semi-supervised learning~\cite{chapelle2009semi,gao2019note,zhu2009introduction}.
As a result of the annotation insufficiency in each image, cross-dataset training encounters the problem of confusing positive and negative samples during sample selection.

Recent studies focus on generating missing annotations through detection submodels trained using each dataset individually to circumvent this difficulty.
Continuing with the previous example, in the studies conducted by ~\cite{rame2018omnia,Zhao_UniDet_ECCV20}, the missing ``\textit{car}'' annotation is predicted on datasets Beta and Gamma through the submodel trained using dataset Alpha. 
We regard the approach employed in these studies as a static supervisor framework because \textit{the submodel only looks once}.
We can only hope that the submodel is strong enough to generate optimal annotations, i.e., both high recall and high precision. 
However, in practice, owing to the capability discrepancy across different models and categories, the quality of annotation generation cannot be guaranteed.
There is no doubt that cross-dataset training for object detection still requires further exploration.

In this study, we first discuss the difference in the quality of annotation generation between hard-label training and soft-label training.
Although the detection performance of both approaches is similar, the annotations generated through hard-label training tend to have higher confidence scores and more false positives than those generated through soft-label training.
Therefore, we conclude that hard-label training is conducive to generating high-recall annotations, whereas soft-label training obtains high-precision annotations.
Inspired by the above, we propose a dynamic supervisor framework for cross-dataset training.
It leverages both the submodel from hard-label training and that from soft-label training to expand and shrink the generated annotations dynamically and obtain a final annotation set with high recall and high precision.
In other words, the proposed dynamic supervisor updates the annotations multiple times according to the multiple-updated submodels, avoiding the dilemma of \textit{only looking once}.
As shown in Figure~\ref{fig:fig1}, the annotations (highlighted in blue) generated using the initial submodel only cover a portion of the \textit{car} instances.
The hard-label-training submodel then adds additional predictions (highlighted in yellow) to increase the recall rate.
Afterward, the soft-label-training submodel re-checks the annotations and reduces the unreliable predictions (the blue dotted rectangle) to increase the precision rate.
In this study, we reveal the implicit connections between hard- and soft-label training and the methods for pseudo-annotation ensembling.
The dynamic supervisor framework is designed to utilize these connections and achieve superior performance.
Extensive experiments conducted on three combinations of several existing datasets~\cite{everingham2010pascal,lin2014microsoft,song2015sun,mogelmose2012vision} support our analyses and proposed method.
The main contributions of this study are as follows:
\begin{enumerate}
\item We show that hard-label training and soft-label training are conducive to improving the recall and precision of predictions, respectively, and that their integration brings improvements to both.
\item We propose a dynamic supervisor framework, which dynamically polishes the annotations and adaptively selects predictions based on category.
\item We conduct detailed experiments to demonstrate the effectiveness of the proposed framework, and we achieve state-of-the-art performance on all three cross-dataset settings.
\end{enumerate}

\section{Related Work}
Over the past decade, cross-dataset learning has become increasingly popular.
In object detection, cross-dataset learning is significantly different from image classification or image retrieval tasks.
When multiple datasets annotated with different category sets are merged, the incomplete annotation transforms fully supervised learning into semi-supervised learning~\cite{radosavovic2018data,jeong2019consistency,wang2018cost}.
Radosavovic et al.~\cite{radosavovic2018data} proposed a method called data distillation, which ensembles predictions from multiple transformations of unlabeled data, using a single model to automatically generate new training annotations.
Jeong et al.~\cite{jeong2019consistency} proposed consistency constraints to enhance detection performance.
Wu et al.~\cite{wu2018soft} proposed a soft sampling method that re-weighs the gradients of RoIs as a function of overlap with positive instances.
This method ensures that uncertain background regions are given a smaller weight compared to that of the hard-negatives when there are some unlabeled object instances in training images.
Yang et al.~\cite{yang2020object} proposed treating object detection as a positive-unlabeled problem, which removes the assumption that unlabeled regions must be negative.
Chadwick and Newman~\cite{chadwick2019training} examined the effect of different types of label noise on the performance of an object detector and applied the co-teaching framework to improve the performance of the detector trained on a noisy dataset. 
These works exploited their methods to make full use of the unlabeled or noisy-labeled data, but could not expand the capability boundary of detection models.

Many existing works~\cite{Zhao_UniDet_ECCV20,rame2018omnia,abbasi2020self} attempt to tackle the cross-dataset object detection with pseudo-labeling methods.
Abbasi et al.~\cite{abbasi2020self} proposed a computationally efficient self-supervised framework to create pseudo-labels for the unlabeled positive instances in the merged dataset in order to train the object detector jointly on both ground truths and pseudo labels.
Rame et al.~\cite{rame2018omnia} proposed the use of models trained using several datasets individually to generate pseudo-annotations on other datasets, and they proposed a new classification loss called SoftSig to handle unreliable pseudo-annotations.
Zhao et al.~\cite{Zhao_UniDet_ECCV20} exploited a pseudo-labeling approach and proposed loss functions that carefully integrated partial but correct annotations with complementary but noisy pseudo-labels.

In the training of supervised neural networks, the different usage of labels results in different properties of neural networks.
Soft-label training can enhance the smoothness of output probabilities and prevent overconfident predictions~\cite{zou2019confidence}. 
M{\"u}ller et al.~\cite{Mller2019WhenDL} found that training a network using hard labels typically results in the correct logit being much larger than any of the incorrect logits, and it also allows the incorrect logits to be significantly different from one another.
Lukasik et al.~\cite{lukasik2020does} reported that label smoothing is effective as a technique for coping with label noise, and it improves the accuracy of both the clean and noisy parts of the data.

\section{Analysis of Hard- and Soft-Label Training}
\label{analysis}
We dedicate a separate section to dissect the differences in the quality of annotation generation between hard-label training and soft-label training.
Without loss of generality, we adopt two datasets with different category sets for cross-dataset training.
For the experiments in this section, we choose the single-shot detector (SSD)~\cite{liu2016ssd} to avoid disturbing factors introduced from complex detection frameworks.

\subsection{Experiment Setting} \label{section3-1}
\textbf{Dataset.}
We sample images from the MS COCO dataset~\cite{lin2014microsoft} to establish two mini datasets, namely \textit{miniCOCO-Alpha} and \textit{miniCOCO-Beta}.
Each of them contains 8K images for training and 5K images for validation.
We preserve 10 category labels on \textit{miniCOCO-Alpha}:
\{\textit{car, handbag, truck, light, bench, chair, horse, bicycle, cup, plant}\},
and we preserve an additional 10 category labels on \textit{miniCOCO-Beta}:
\{\textit{person, bottle, motorcycle, bird, boat, umbrella, sheep, wine glass, table, tv}\}.
For simplicity, \textit{miniCOCO-Alpha} is denoted as $(\mathbf{I}^\alpha, \mathbf{C}^\alpha_1)$ and \textit{miniCOCO-Beta} is denoted as $(\mathbf{I}^\beta, \mathbf{C}^\beta_2)$, where $\mathbf{I}$, $\mathbf{C}_1 $ and $\mathbf{C}_2$ denote the image set, the first 10-category annotation set, and the second 10-category annotation set, respectively.
Numbers 1 and 2 correspond to the first 10-category and the second 10-category, respectively.
This setting models the central issue when multiple datasets with different category sets are merged.

As we discussed in Section~\ref{introduction}, the key step in cross-dataset training for object detection involves seeking the accurate annotation sets $\mathbf{C}_2^\alpha$ and $\mathbf{C}_1^\beta$, and thus, transforming the cross-dataset $(\{\mathbf{I}^\alpha, \mathbf{I}^\beta\}, \{\mathbf{C}^\alpha_1, \mathbf{C}^\beta_2\})$ into a complete form $(\{\mathbf{I}^\alpha, \mathbf{I}^\beta\}, \{\mathbf{C}^\alpha_1, \mathbf{C}^\alpha_2, \mathbf{C}^\beta_1, \mathbf{C}^\beta_2\})$.
Next, we conduct a detailed experimental analysis on the generation of the missing annotation sets, $\mathbf{C}_2^\alpha$ and $\mathbf{C}_1^\beta$, with high quality.

\textbf{Training Details.}
ResNet-50~\cite{he2016deep} pretrained on ImageNet~\cite{russakovsky2015imagenet} is chosen as the backbone of the SSD in this section.
During training, we use five image scales \{448, 480, 512, 544, 576\} (the aspect ratio of the image is 1:1) randomly.
The network is trained using the stochastic gradient descent (SGD) algorithm for 100 epochs with 0.9 momentum, 0.0005 weight decay, and 32 batch sizes on two NVIDIA V100 GPUs.
The initial learning rate is 0.0026 and it decays at epochs 66 and 83.
The loss functions used during training are the cross-entropy loss for the classification branch and the smooth-L1 loss for the regression branch.

\textbf{Inference Details.}
During the inference phase, the input image is resized to 576 $\times$ 576, after which it is input into the entire network to output the predicted bounding boxes with a predicted category.
To filter out the redundant background bounding boxes, the confidence threshold is set to 0.01, and non-maximum suppression (NMS) is applied, with an IoU threshold of 0.6 per class, when evaluating the detection performance of the network.

\begin{figure}[t]
\begin{center}
   \includegraphics[width=0.85\linewidth]{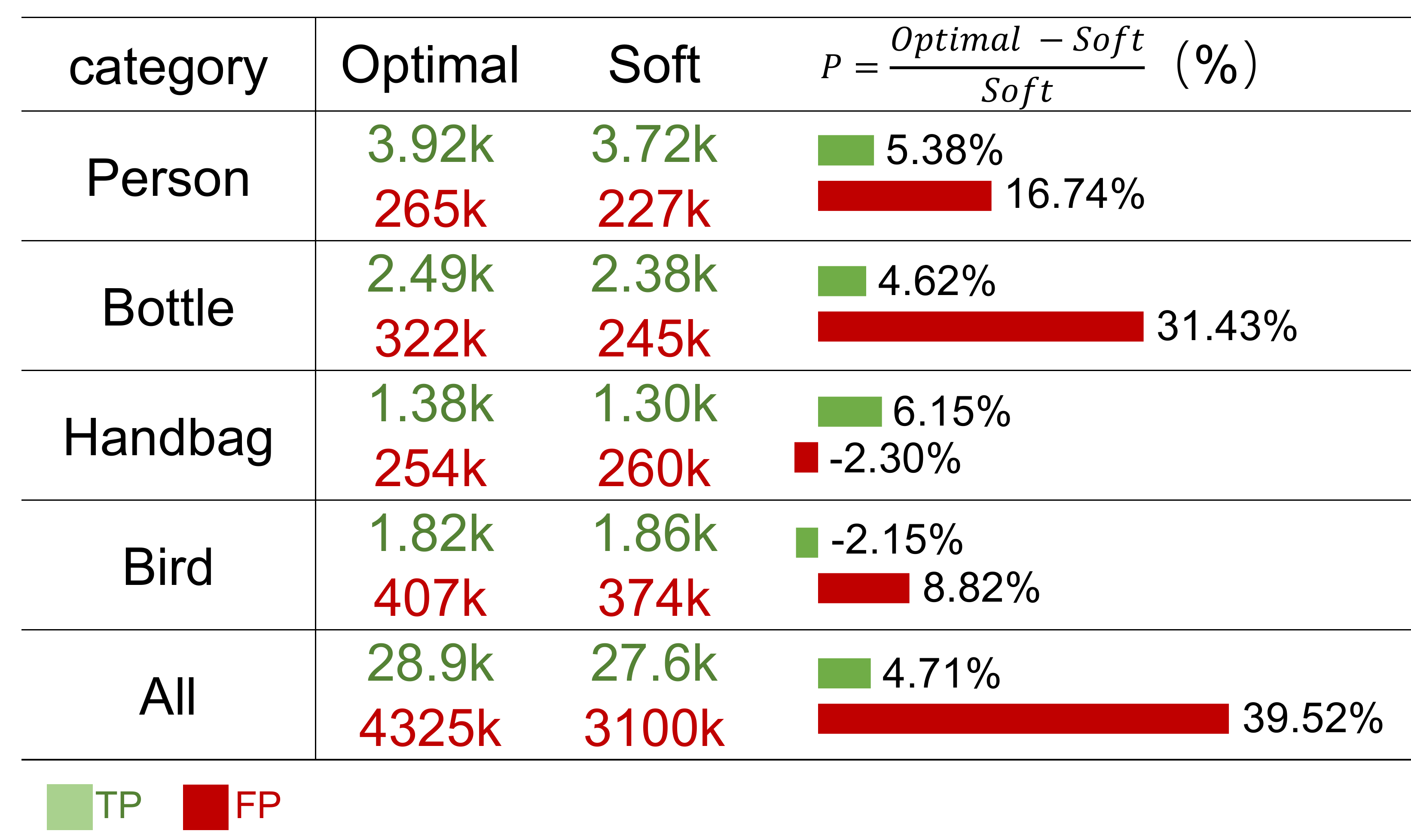}
\end{center}
   \caption{TP (in green color) and FP (in red color) quantities of detection outputs for the theoretically optimal model and soft-labeling-training model.
   The statistics of 4 specific categories are listed here.
   \textit{``All''} denotes the total quantities of 20 categories.
   $P$ reflects that the TP quantities of two models are similar but the FP quantity of the theoretically optimal model is much more than that of the soft-labeling-training model.
}
\label{fig:tp-num}
\end{figure}

\subsection{Statistical Characteristic}
A submodel trained using \textit{miniCOCO-Alpha} generates the first 10-category annotation set for \textit{miniCOCO-Beta}, which is denoted as $\mathbf{iC}_1^\beta$.
Correspondingly, another submodel trained using \textit{miniCOCO-Beta} generates the second 10-category annotation set for \textit{miniCOCO-Alpha}, which is denoted as $\mathbf{iC}_2^\alpha$.
It is worth noting that each element in $\mathbf{iC}_2^\alpha$ and $\mathbf{iC}_1^\beta$ consists of bounding box coordinates, a category label, and the corresponding confidence score.
Additionally, the confidence score of each element should be larger than the threshold $T_c$.
The next conventional operation involves integrating $\mathbf{iC}_2^\alpha$ and $\mathbf{iC}_1^\beta$ into the datasets and establishing a complete one: $(\{\mathbf{I}^\alpha, \mathbf{I}^\beta\}, \{\mathbf{C}^\alpha_1, \mathbf{iC}^\alpha_2, \mathbf{iC}^\beta_1, \mathbf{C}^\beta_2\})$.
Soft-label training is the usual choice for this type of ``noisy-annotated'' dataset. 
During training, the loss function for the classification branch is expressed as follows: 
$L_{cls} = -\sum_{c=0}^{20} y_c\log p_c$.
The label for a pseudo-annotation of category $j$ and confidence $s$ is a soft form ${\boldsymbol {y} } = \left [ y_0, y_1,...,y_{20} \right ]^{T}$, where $y_0 = 1 - s$ and $y_j = s$.
Through the dataset $(\{\mathbf{I}^\alpha, \mathbf{I}^\beta\}, \{\mathbf{C}^\alpha_1, \mathbf{iC}^\alpha_2, \mathbf{iC}^\beta_1, \mathbf{C}^\beta_2\})$, the soft-label-training submodel achieves 34.4$\%$ mAP on the validation set (5K images from \textit{miniCOCO-Alpha} and 5K images from \textit{miniCOCO-Beta}).

Because \textit{miniCOCO-Alpha} and \textit{miniCOCO-Beta} are both sampled from MS COCO, we can obtain the true annotations directly, namely $\mathbf{tC}_2^\alpha$ and $\mathbf{tC}_1^\beta$, where $\mathbf{t}$ denotes true.
Using the dataset $(\{\mathbf{I}^\alpha, \mathbf{I}^\beta\}, \{\mathbf{C}^\alpha_1, \mathbf{tC}^\alpha_2, \mathbf{tC}^\beta_1, \mathbf{C}^\beta_2\})$, a theoretically optimal SSD model is trained, and it achieves 38.2$\%$ mAP on the same validation set.

We then analyze the predictions of both models on the validation set.
As shown in Figure~\ref{fig:tp-num}, the statistical characteristics have attracted our attention.
The quantity of true positives (TP) from the soft-label-training model is similar to that from the theoretically optimal model (the quantity from the theoretically optimal model is approximately 5\% higher than that from the soft-label-training model), which means that the maximum recall between the two models is close. %and the maximum recall for the soft-label-training model is slightly lower.
However, when we observe the quantities of false positives (FP), we find that the theoretically optimal model outputs much more FP than the soft-label-training model, which is counter-intuitive.

\begin{figure}[t]
\begin{center}
   \includegraphics[width=1\linewidth]{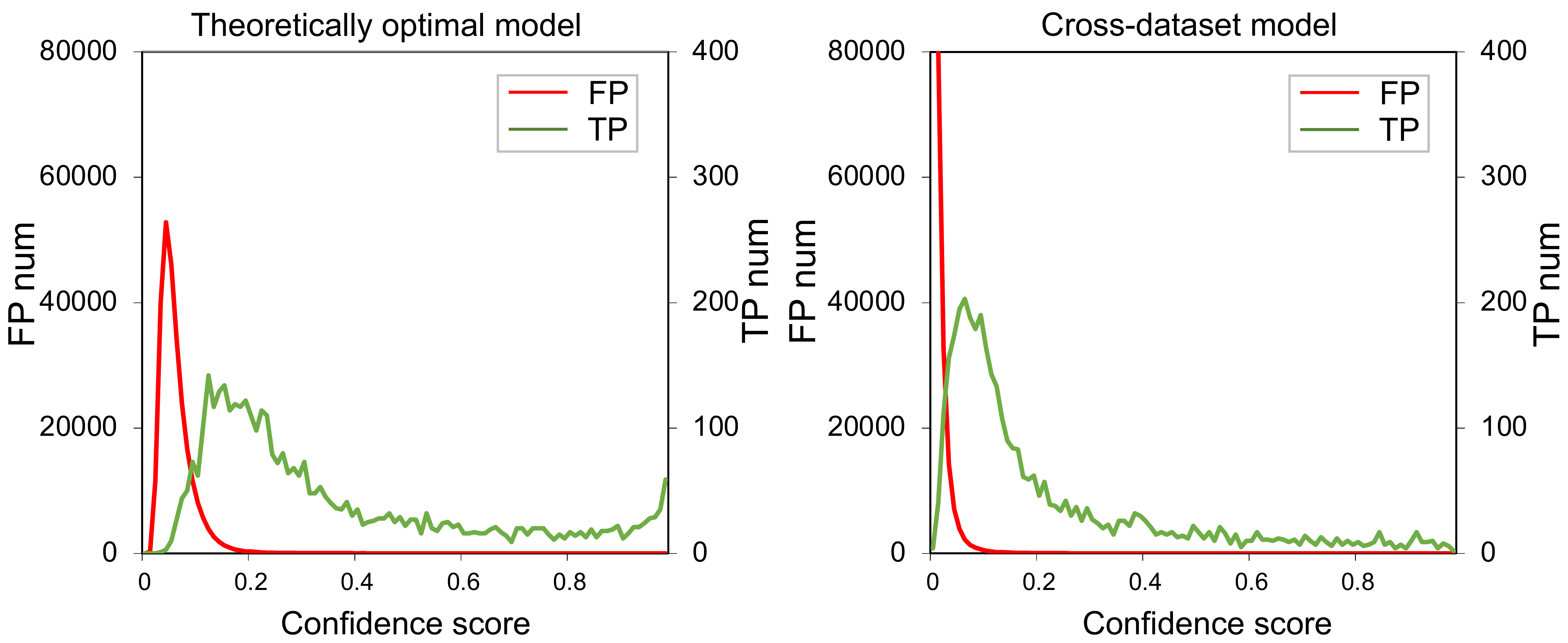}
\end{center}
   \caption{The numbers of detection results with different confidence scores of TP and FP respectively.
   \textbf{Left}: detection output distribution of the theoretically optimal model.
   \textbf{Right}: detection output distribution of the cross-dataset model (soft-label training). 
}
\label{fig:tp-fp}
\end{figure}

\begin{figure}[t]
\begin{center}
   \includegraphics[width=0.85\linewidth]{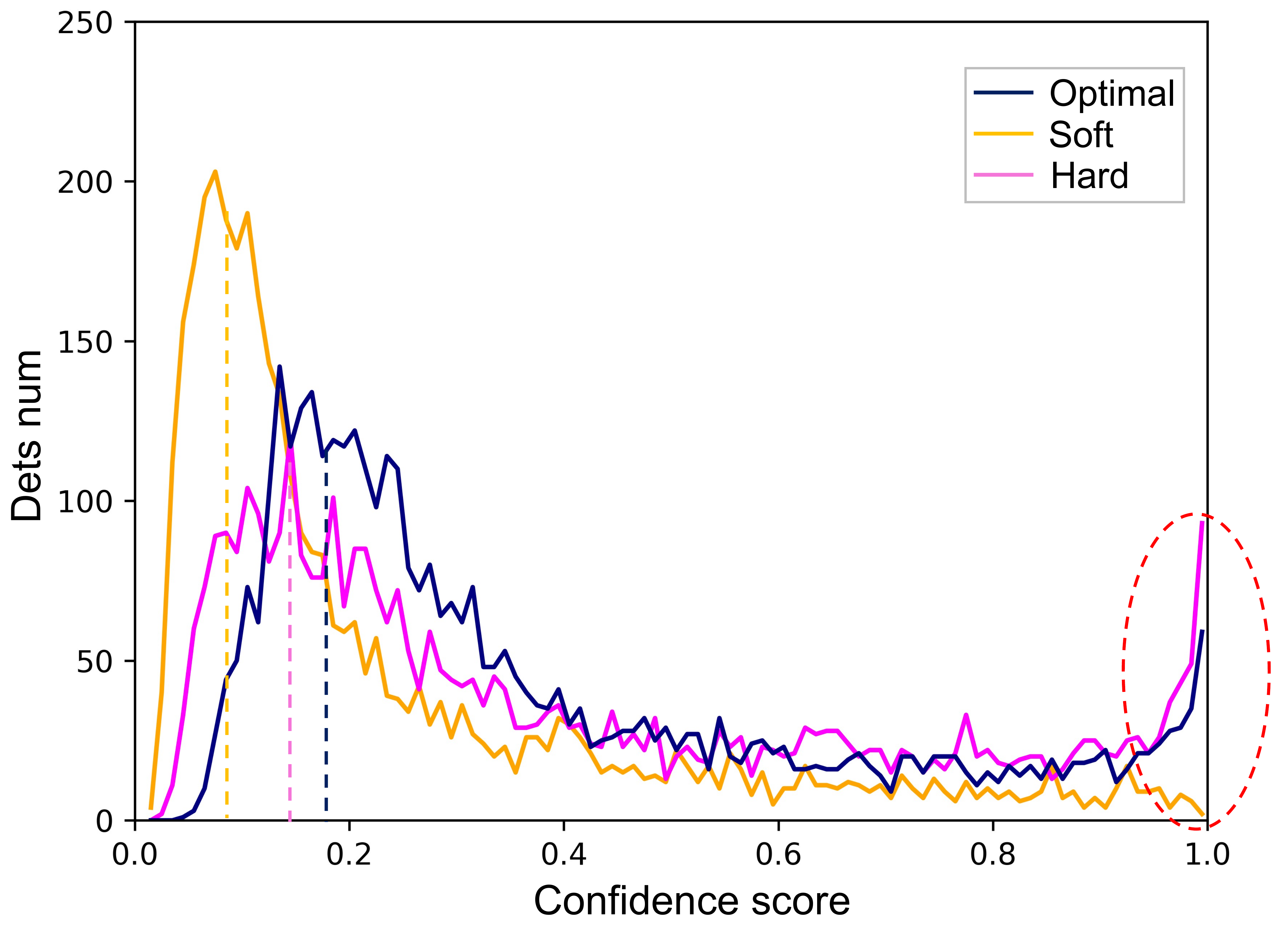}
\end{center}
   \caption{The numbers of TP with different confidence scores for three models.
   \textit{``Optimal''} is the curve of the theoretically optimal model.
   \textit{``Soft''} and \textit{``Hard''} is the curve of the soft-label-training model and hard-label-training model respectively.
}
\label{fig:tp-three}
\end{figure}

We suggest that the performance gap between the two models is caused by the distribution of detection predictions rather than the quantity of detection outputs.
Figure~\ref{fig:tp-fp} shows the output distribution of category \textit{person}, where the figure on the left is for the theoretically optimal model and the one on the right is for the soft-label-training model.
The TP from the theoretically optimal model tend to have a higher confidence score than those of the soft-label-training model.
In contrast, more TP from the soft-label-training model are concentrated in the low-score region and severely mixed up with FP, which should be the direct cause of the performance gap between the two detectors.
Quantitatively, the TP of the theoretically optimal model that have a confidence score higher than 0.2 account for 68.2\% of all TP, whereas those of the soft-label-training model account for a mere 36.3\%.
As for FP with a confidence score higher than 0.2, their number in the theoretically optimal model is almost four times that in the soft-label-training model.

To verify the difference between the two models, we conduct hard-label training on the dataset $(\{\mathbf{I}^\alpha, \mathbf{I}^\beta\}, \{\mathbf{C}^\alpha_1, \mathbf{iC}^\alpha_2, \mathbf{iC}^\beta_1, \mathbf{C}^\beta_2\})$ for further analysis.
The true positive distributions of category \textit{person} for the theoretically optimal model, soft-label-training model, and hard-label-training model are illustrated in Figure~\ref{fig:tp-three}.
The distribution curves of the theoretically optimal model and the hard-label-training model are similar.
They output more TP in the high-score region, and the peaks of their curves are more to the right than the peak of the soft-label-training model’s curve.
The TP of the hard-label-training model that have a confidence score higher than 0.2 account for 64.7\% of all TP, which is close to that of the theoretically optimal model (68.2\%).
Additionally, the high-scoring (\textgreater 0.9) TP account for 6.8\% and 9.5\% for the theoretically optimal model and the hard-label-training model, respectively, which is higher than that of the soft-label-training model (2.1\%).
Based on these statistics, we suggest that the prediction of the hard-label-training model (the theoretically optimal model is also trained using hard labels) has a higher mean score and contains more FP, whereas that of the soft-label-training model has a lower mean score and contains fewer FP.

\begin{figure*}
\begin{center}
  \includegraphics[width=1\linewidth]{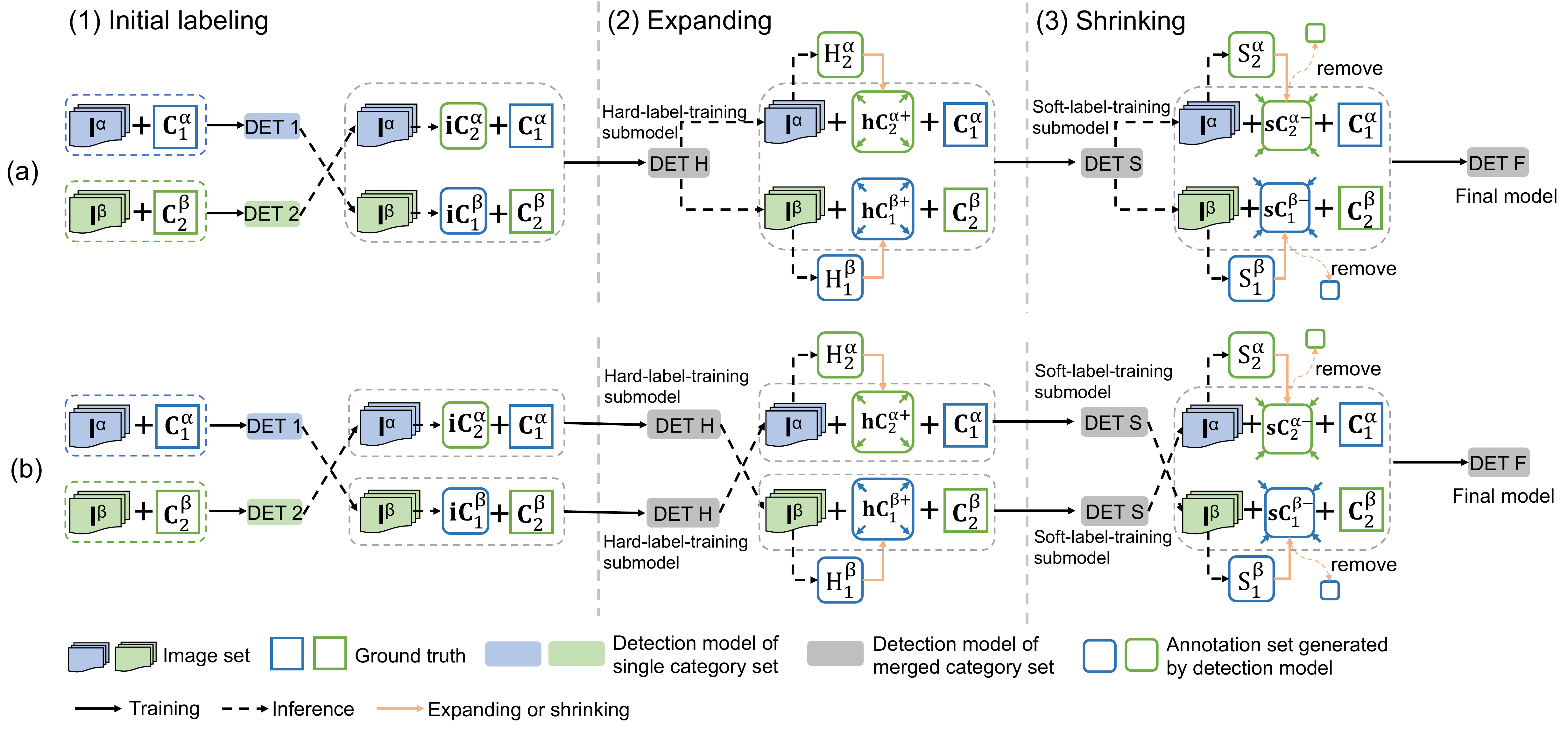}
\end{center}
   \caption{The structure of dynamic supervisor framework.
   (a) Self-annotated mechanism.
   (b) Cross-annotated mechanism.
   The notations in Section~\ref{analysis} are followed here.
   $\alpha$ and $\beta$ correspond to different image sets and numbers 1 and 2 correspond to different category sets.}
\label{fig:framework}
\end{figure*}

\subsection{Update on Supervisor}
Considering the different characteristics of hard-label training and soft-label training, we propose taking advantage of these two strategies to improve the quality of annotations $\mathbf{iC}_2^\alpha$ and $\mathbf{iC}_1^\beta$ from two perspectives.
1) We can \textit{expand} two annotation sets by adding new predictions that can increase the recall of annotations.
2) We can \textit{shrink} two annotation sets by filtering out unreliable predictions to improve the precision of annotations.

Specifically, images from $\mathbf{I^\alpha}$ and $\mathbf{I^\beta}$ are input into the hard-label-training model and the soft-label-training model to obtain new detection results, $\left ( \mathbf{H}_2^\alpha, \mathbf{H}_1^\beta \right )$ and $\left ( \mathbf{S}_2^\alpha, \mathbf{S}_1^\beta \right )$, respectively. $\mathbf{H}$ denotes the predictions produced using the hard-label-training model, and $\mathbf{S}$ denotes the predictions produced using the soft-label-training model.
Therefore the expanding operation using hard-label training is formulated as follows:
\begin{equation}
   \begin{cases}
    \mathbf{hC}_2^{\alpha+} = \mathbf{iC}_2^\alpha \cup (\mathbf{H}_2^\alpha \setminus (\mathbf{C}_1^\alpha \cup \mathbf{iC}_2^\alpha))\\
    \mathbf{hC}_1^{\beta+} = \mathbf{iC}_1^\beta \cup (\mathbf{H}_1^\beta \setminus (\mathbf{C}_2^\beta \cup \mathbf{iC}_1^\beta))
   \end{cases}
   \label{hard-expand}
\end{equation}
where $\mathbf{hC}_2^{\alpha+}$ denotes the expanded annotation set on dataset \textit{miniCOCO-Alpha} and $\mathbf{hC}_1^{\beta+}$ denotes that on \textit{miniCOCO-Beta}.
The operation $\setminus$ minuses the annotation in $\mathbf{H}_2^\alpha$, which has an IoU larger than a threshold $T_e$, with any annotation in $(\mathbf{C}_1^\alpha \cup \mathbf{iC}_2^\alpha)$.
Similarly, the expanding operation using soft-label training is formulated as follows:
\begin{equation}
   \begin{cases}
    \mathbf{sC}_2^{\alpha+} = \mathbf{iC}_2^\alpha \cup (\mathbf{S}_2^\alpha \setminus (\mathbf{C}_1^\alpha \cup \mathbf{iC}_2^\alpha))\\
    \mathbf{sC}_1^{\beta+} = \mathbf{iC}_1^\beta \cup (\mathbf{S}_1^\beta \setminus (\mathbf{C}_2^\beta \cup \mathbf{iC}_1^\beta))
    \end{cases}
\end{equation}

We formulate the shrinking operation using hard-label training as follows:
\begin{equation}
   \begin{cases}
   \mathbf{hC}_2^{\alpha-} = \mathbf{iC}_2^\alpha \cap (\mathbf{H}_2^\alpha \setminus \mathbf{C}_1^\alpha)\\
   \mathbf{hC}_1^{\beta-} = \mathbf{iC}_1^\beta \cap (\mathbf{H}_1^\beta \setminus \mathbf{C}_2^\beta)
   \end{cases}
\end{equation}
where $\mathbf{hC}_2^{\alpha-}$ denotes the shrunk annotation set on dataset \textit{miniCOCO-Alpha} and $\mathbf{hC}_1^{\beta-}$ denotes that on \textit{miniCOCO-Beta}.
The operation $\cap$ preserves the annotation in $\mathbf{iC}_2^\alpha$, which has an IoU larger than a threshold $T_s$, with any annotation in $(\mathbf{H}_2^\alpha \setminus \mathbf{C}_1^\alpha)$.
It is inspired by the intuition that the bounding boxes detected using both detectors are more reliable.
Similarly, the shrinking operation using soft-label training is formulated as follows:
\begin{equation}
   \begin{cases}
   \mathbf{sC}_2^{\alpha-} = \mathbf{iC}_2^\alpha \cap (\mathbf{S}_2^\alpha \setminus \mathbf{C}_1^\alpha)\\
   \mathbf{sC}_1^{\beta-} = \mathbf{iC}_1^\beta \cap (\mathbf{S}_1^\beta \setminus \mathbf{C}_2^\beta)
   \end{cases}
   \label{soft-shrink}
\end{equation}

\begin{table}[t]
\begin{center}
\begin{tabular}{l | c c |c}
\hline
 Annotations&$\Delta $Recall(\%)&$\Delta $Precision(\%)&mAP(\%)\\
\hline
$\mathbf{iC}^\alpha_2$, $\mathbf{iC}^\beta_1$            &0&0&34.4\\
\hline\hline
$\mathbf{hC}_2^{\alpha+}, \mathbf{hC}_1^{\beta+}$    &\textbf{\textcolor[RGB]{2,167,71}{+18.8}}&\textcolor{red}{-11.8}&\textbf{34.8}\\
$\mathbf{sC}_2^{\alpha+}, \mathbf{sC}_1^{\beta+}$    &\textcolor[RGB]{2,167,71}{+9.1}&\textcolor{red}{-0.7}&34.5\\
\hline\hline
$\mathbf{hC}_2^{\alpha-}, \mathbf{hC}_1^{\beta-}$    &\textcolor{red}{-5.1}&\textcolor[RGB]{2,167,71}{+12.8}&33.7\\
$\mathbf{sC}_2^{\alpha-}, \mathbf{sC}_1^{\beta-}$    &\textcolor{red}{-11.6}&\textbf{\textcolor[RGB]{2,167,71}{+27.6}}&34.4\\
\hline
\end{tabular}
\end{center}
\caption{The relative variation of recall and precision when models expand or shrink the initial pseudo annotation sets ($T_e=0.6$, $T_s=0.6$).
Performance of detection models trained on different pseudo annotation sets.
``$+$'' and ``$-$'' denote expanding and shrinking operations, respectively.
``\textbf{h}'' and ``\textbf{s}'' denote hard-label training and soft-label training, respectively.
}
\label{quality-change}
\end{table}

Based on the true annotations $\mathbf{tC}_2^\alpha$ and $\mathbf{tC}_1^\beta$, we can obtain the relative variation of recall and precision for new annotation sets when we expand or shrink them using different models.
The results are listed in Table \ref{quality-change}, and they confirm our conclusion that the hard-label-training model is conducive to improving the recall of annotation sets, whereas the soft-label-training model is useful for improving the precision of annotation sets.
The performance of the models trained using different annotation sets is also shown in Table \ref{quality-change}.

\section{Dynamic Supervisor}
According to the analysis conducted in Section~\ref{analysis}, a single operation (expanding or shrinking) on the annotation set cannot improve its recall and precision simultaneously.
Inspired by the characteristics of hard-label training and soft-label training, we delve into a method for improving the quality of the annotation set, and we propose a dynamic supervisor framework to produce a more complete annotation set progressively.

\subsection{The Overall Framework} \label{section4-1}
The structure of the proposed dynamic supervisor framework is shown in Figure~\ref{fig:framework}, and two datasets annotated with different category sets are merged in our illustration (the notations in Section~\ref{analysis} are followed).
There are three steps for generating the final annotation set.
The first step of the dynamic supervisor framework is similar to the one employed in previous studies~\cite{rame2018omnia,Zhao_UniDet_ECCV20}, where two detection models are trained using \textit{miniCOCO-Alpha} or \textit{miniCOCO-Beta} individually, after which each detection model generates an initial annotation set for images from the other dataset (\textit{Initial labeling}, as shown in Figure~\ref{fig:framework}).

By combining the ground truths ($\mathbf{C}_1^\alpha$, $\mathbf{C}_2^\beta$) with the generated annotation sets ($\mathbf{iC}_2^\alpha$, $\mathbf{iC}_1^\beta$), \textit{miniCOCO-Alpha} and \textit{miniCOCO-Beta} can be merged.
One optional structure of the dynamic supervisor framework is shown in Figure~\ref{fig:framework} (a), where two datasets are merged after the generation of the initial annotation sets.
Next, in the second step (\textit{Expanding} in Figure~\ref{fig:framework}), a detection model trained using $(\{\mathbf{I}^\alpha, \mathbf{I}^\beta\}, \{\mathbf{C}^\alpha_1, \mathbf{iC}^\alpha_2, \mathbf{iC}^\beta_1, \mathbf{C}^\beta_2\})$ generates new annotations through hard-label training to expand the initial annotation sets ($\mathbf{iC}_2^\alpha$, $\mathbf{iC}_1^\beta$) into ($\mathbf{hC}_2^{\alpha+}$, $\mathbf{hC}_1^{\beta+}$).
In the third step (\textit{Shrinking} in Figure~\ref{fig:framework}), another detection model trained using $(\{\mathbf{I}^\alpha, \mathbf{I}^\beta\}, \{\mathbf{C}^\alpha_1, \mathbf{hC}_2^{\alpha+}, \mathbf{hC}_1^{\beta+}, \mathbf{C}^\beta_2\})$ filters out unreliable annotations through soft-label training to shrink the expanded annotation sets ($\mathbf{hC}_2^{\alpha+}$, $\mathbf{hC}_1^{\beta+}$) into ($\mathbf{sC}_2^{\alpha-}$, $\mathbf{sC}_1^{\beta-}$).
Based on the dataset $(\{\mathbf{I}^\alpha, \mathbf{I}^\beta\}, \{\mathbf{C}^\alpha_1, \mathbf{sC}_2^{\alpha-}, \mathbf{sC}_1^{\beta-}, \mathbf{C}^\beta_2\})$, we can obtain the final model for cross-dataset object detection.

In the first structure of the dynamic supervisor discussed above, the detection models (in the second and third steps) must update their supervision information independently.
Considering the risk that the models are prone to overfitting to noisy annotations in this ``self-annotated mechanism'', we propose another structure, which is a ``cross-annotated mechanism,'' as shown in Figure~\ref{fig:framework} (b).
Unlike in the first structure, the two datasets are not merged after the generation of the initial annotation sets in the first step.
When the two original datasets are augmented with the generated annotation sets $(\mathbf{I}^\alpha, \mathbf{C}^\alpha_1, \mathbf{iC}^\alpha_2)$ and $(\mathbf{I}^\beta, \mathbf{iC}^\beta_1, \mathbf{C}^\beta_2)$, they are responsible for training two models individually through hard-label training.
The two hard-label-training models then generate new annotations for images from the other dataset to expand their initial annotation set in the second step.
Because these two models are trained using the augmented datasets, their detection distributions are different from those of the previous two models in the first step.
Therefore, they can detect objects that are ignored, thereby improving the recall of the annotation sets.
Although the recall is improved, the expanding operation probably results in a decrease in precision.
Accordingly, two other models are then trained through soft-label training using the expanded annotation sets ($(\mathbf{I}^\alpha, \mathbf{C}^\alpha_1, \mathbf{hC}_2^{\alpha+})$ or $(\mathbf{I}^\beta, \mathbf{hC}_1^{\beta+}, \mathbf{C}^\beta_2)$).
The new annotations generated using two soft-label-training models are used to filter out the unreliable annotations of the expanded annotation sets in the third step.
The objects detected through the two different models are more likely to be TP than FP. Therefore, there will be an improvement in the precision of annotation sets after the shrinking operation.

Finally, \textit{miniCOCO-Alpha} and \textit{miniCOCO-Beta} are merged, and the final model is trained using this multiple-updated dataset: $(\{\mathbf{I}^\alpha, \mathbf{I}^\beta\}, \{\mathbf{C}^\alpha_1, \mathbf{sC}_2^{\alpha-}, \mathbf{sC}_1^{\beta-}, \mathbf{C}^\beta_2\})$.
We suppose that the recall and precision of the annotation sets can be promoted comprehensively through the multiple update steps.

\subsection{Experimental Verification}

\textbf{Cross-dataset Setting.}
To verify the effectiveness of the proposed dynamic supervisor framework, we define three different combinations of datasets that cover diverse scenarios.
In Table~\ref{dataset-setting}, we summarize the dataset settings used in our experiments.
Setting A contains two mini datasets sampled from MS COCO, which we used in Section~\ref{analysis}.
Setting B combines PASCAL VOC~\cite{everingham2010pascal} (20 categories) and MS COCO~\cite{lin2014microsoft} (without VOC categories) to verify the performance of the proposed method on large-scale datasets.
Setting C consists of three datasets from different scenarios: PASCAL VOC is a general dataset containing 20 common categories, SUN-RGBD~\cite{song2015sun} includes 18 categories of indoor scenes, and LISA-Signs~\cite{mogelmose2012vision} is a driving scenes dataset, which contains 4 traffic signs.
This setting combines multiple datasets with large gaps, and frequent scene overlapping exists among these datasets.

\begin{table}[t]
\small
\begin{center}
\begin{tabular}{c | c }
\hline
 Setting&Datasets\\
\hline\hline
A&miniCOCO-Alpha~\cite{lin2014microsoft} $+$ miniCOCO-Beta~\cite{lin2014microsoft}\\
B&VOC~\cite{everingham2010pascal} $+$ COCO~\cite{lin2014microsoft} (w/o VOC categories)\\
C&VOC~\cite{everingham2010pascal} $+$ SUN-RGBD~\cite{song2015sun} $+$ LISA-Signs~\cite{mogelmose2012vision}\\
\hline
\end{tabular}
\end{center}
\caption{Three combinations of datasets in our experiments. Setting A is the dataset combination in Section~\ref{analysis}.
B and C are the same setting as that of~\cite{Zhao_UniDet_ECCV20}.
}
\label{dataset-setting}
\end{table}

\begin{table}[t]
\begin{center}
\begin{tabular}{c | c c}
\hline
 Method&COCO&MIX\\
\hline\hline
Naive-combination&42.6&43.7\\
Partial-loss~\cite{cour2011learning}&43.6&44.6\\
UOD~\cite{wang2019towards} + Merge&45.6&46.1\\
Pseudo-Labeling~\cite{Zhao_UniDet_ECCV20}&50.3&52.2\\
Static Supervisor (ours)&53.3&51.5\\
Dynamic Supervisor (ours)&\textbf{56.2}&\textbf{55.8}\\
\hline
\end{tabular}
\end{center}
\caption{Detection performance(mAP in \%) of setting B on different validation sets.
The backbone of all methods is ResNet-50.
}
\label{setting-b-result}
\end{table}

\begin{table}[t]
\begin{center}
\begin{tabular}{c | c| c}
\hline
 Method&backbone&mAP\\
\hline\hline
Naive-combination&ResNet-50&58.1\\
Partial-loss~\cite{cour2011learning}&ResNet-50&58.3\\
UOD~\cite{wang2019towards} + Merge&ResNet-50&59.3\\
Min-Entropy loss~\cite{Zhao_UniDet_ECCV20}&ResNet-50&58.7\\
Pseudo-Labeling~\cite{Zhao_UniDet_ECCV20}&ResNet-50&61.1\\
Static Supervisor (ours)&ResNet-50&60.7\\
Dynamic Supervisor (ours)&ResNet-50&\textbf{61.7}\\
\hline
\end{tabular}
\end{center}
\caption{Detection performance(mAP in \%) on setting C.
}
\label{setting-c-result}
\end{table}

\textbf{Comparison with Other Methods.}
We apply the proposed dynamic supervisor framework to the Faster-RCNN~\cite{ren2015faster} model.
ResNet-50~\cite{he2016deep} with FPN~\cite{lin2017feature} is used as the backbone, and it is pretrained on ImageNet~\cite{russakovsky2015imagenet}.
We follow the implementation presented by ~\cite{wu2019detectron2}, where input images are resized to keep their shorter side as one of $\left \{640, 672, 704, 736, 768, 800 \right \}$ randomly and their longer side less than or equal to 1,333 during training.

For setting B, we evaluate our model on two datasets: COCO and MIX~\cite{Zhao_UniDet_ECCV20} (a combination of VOC and COCO).
The experimental results are shown in Table~\ref{setting-b-result}.
Compared with the naive combination of the two datasets (the first row in Table~\ref{setting-b-result}), the partial loss method proposed by ~\cite{cour2011learning} only takes a small step forward.
The following pseudo-labeling method proposed by ~\cite{Zhao_UniDet_ECCV20} achieves much better performance (50.3\% \textit{vs.} 42.6\%, 52.2\% \textit{vs.} 43.7\%), which demonstrates the effectiveness of pseudo-annotation.
The performance of the static supervisor is comparable to that of the one proposed by ~\cite{Zhao_UniDet_ECCV20} (53.3\% \textit{vs.} 50.3\%, 51.5\% \textit{vs.} 52.2\%).
However, the dynamic supervisor framework goes further and achieves a new state-of-the-art performance (56.2\% on COCO and 55.8\% on MIX).
This performance demonstrates the effectiveness of the dynamic supervisor framework in a large-scale dataset setting.

For setting C, there are four overlapping categories between PASCAL VOC and SUN-RGBD. Therefore, the entire merged dataset has 38 categories in total.
A validation set~\cite{Zhao_UniDet_ECCV20} of 1,500 images (from three datasets) is annotated for all 38 categories for evaluation.
The results of the experiments conducted in this setting are listed in Table~\ref{setting-c-result}.
The static supervisor framework achieves a performance that is comparable to that of the pseudo-labeling method~\cite{Zhao_UniDet_ECCV20} (60.7\% \textit{vs.} 61.1\%).
Based on the static supervisor framework, the proposed dynamic supervisor framework enhances the performance further by 1\% in mAP (61.7\% \textit{vs.} 60.7\%).
The best performance in this setting demonstrates the generality of our method in a complicated setting (multiple datasets with frequent scene overlapping).

Comparing the promotions brought by the dynamic supervisor in setting B and setting C, the performance superiority in setting B is larger than that in setting C.
This inapparent superiority in setting C results from the slight annotation missing.
The more annotations are lost, the more promotions the dynamic supervisor will bring.
Quantitively, there are 4.8 predicted instances per image for setting B, whereas there are only 1.0 predicted instances per image for setting C.
In other words, the training task in setting C is closer to a fully-supervised problem, resulting in minor performance gaps among all competitors.

\begin{figure}[t]
\begin{center}
   \includegraphics[width=0.9\linewidth]{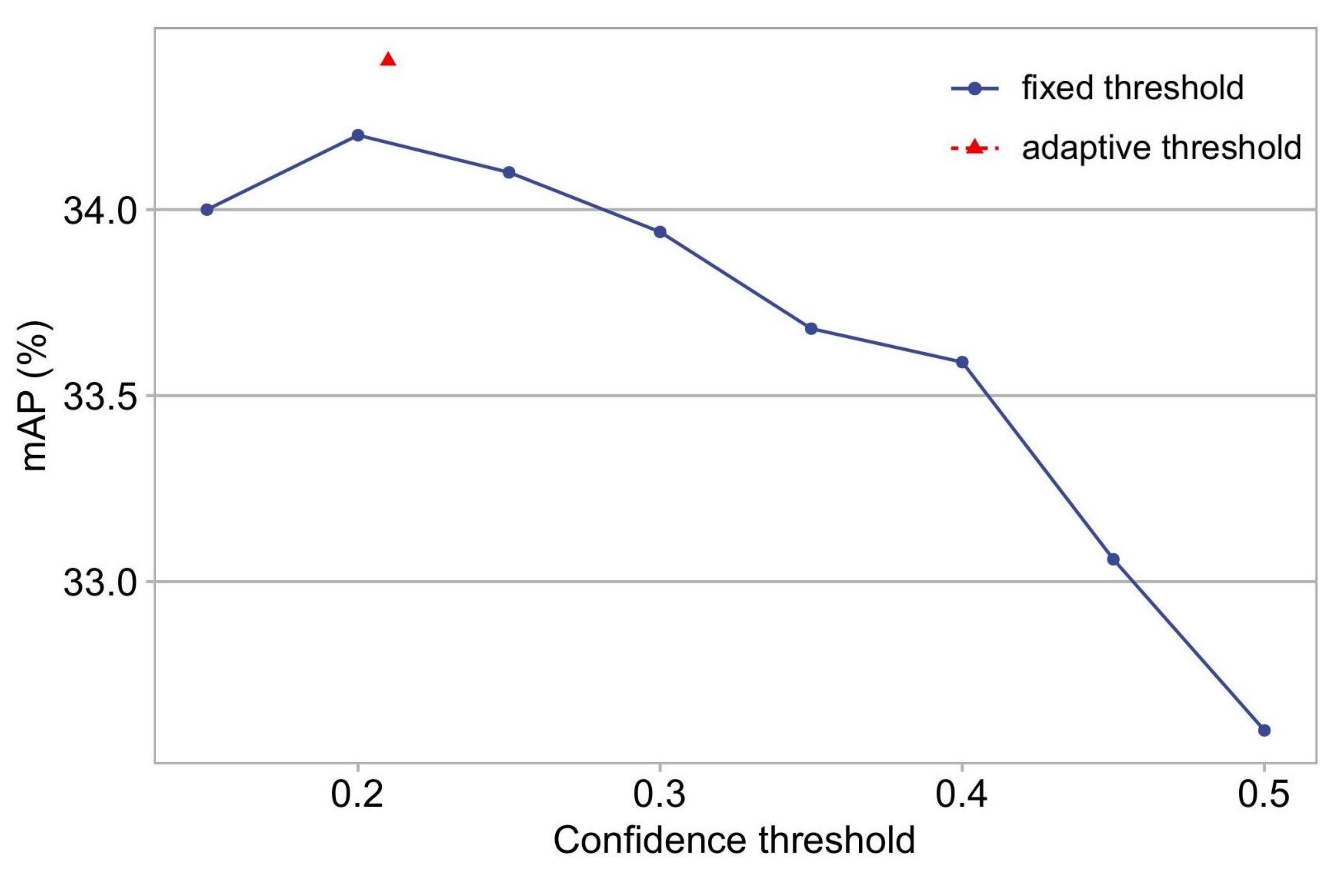}
\end{center}
   \caption{Detection performance (mAP in \%) of models trained on initial annotation sets with fixed confidence threshold and adaptive threshold.
   The average value of adaptive threshold is 0.21 in this setting.
}
\label{fig:confidence}
\end{figure}

\begin{figure}[t]
\begin{center}
   \includegraphics[width=0.9\linewidth]{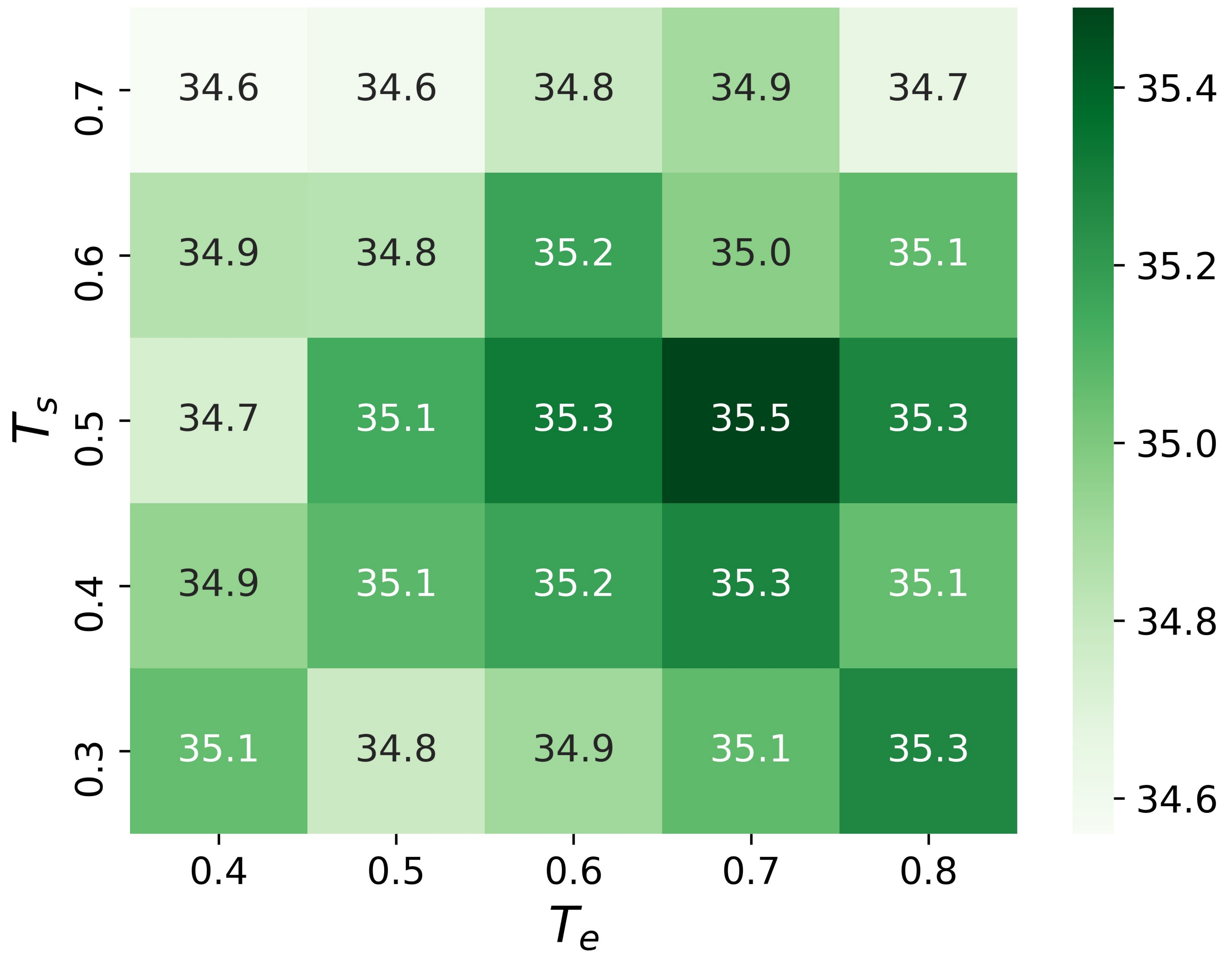}
\end{center}
   \caption{Detection performance (mAP in \%) for different choices of $T_e$ and $T_s$.
}
\label{fig:IoU}
\end{figure}

%-----------------------------------------------------------
\subsection{Ablation Studies}
We follow the controlled experimental setting in Section~\ref{section3-1} to analyze the proposed dynamic supervisor framework.
The implementation details are already described in Section~\ref{section3-1}.
Ablation results are shown in Table~\ref{dynamic-supervisor} - Table~\ref{cross-self}, Figure~\ref{fig:confidence}, and Figure~\ref{fig:IoU}.
Details about ablation studies are discussed in the following.

\textbf{Dynamic Hyperparameters:}
Figure~\ref{fig:confidence} shows the detection performance of models trained using initial annotation sets with a different confidence threshold $T_c$.
It is hard for us to determine which threshold is the best (it should be between 0.15 and 0.3 in this setting), especially when encountering other datasets or detection frameworks.
Meanwhile, considering the recognition abilities of a detection model vary from one category to another, we propose selecting the confidence threshold adaptively.
The confidence threshold for each category is determined when the F1-score of this category reaches the maximum on the validation set.
This method achieves the best performance, as shown in Figure~\ref{fig:confidence}, which is simple but effective.

Figure~\ref{fig:IoU} shows the detection performance for different choices of $T_e$ and $T_s$.
All these hyperparameter combinations can improve the detection performance compared with the baseline model (34.4\% mAP).
From Eq. (\ref{hard-expand}) and (\ref{soft-shrink}), we can observe that a larger $T_e$ introduces more new pseudo-annotations in the expanding operation.
Additionally, a smaller $T_s$ preserves more old pseudo-annotations in the shrinking operation.
As shown in Figure~\ref{fig:IoU}, it is better to apply a large $T_e$ to obtain more recall improvement and a small $T_s$ to avoid many true annotations being filtered out.
In the remainder of this paper, we use $T_e=0.7$ and $T_s=0.5$ for ablation studies.

\begin{table}[t]
\begin{center}
\begin{tabular}{c| c c c |c}
\hline
Method&Pseudo-label&Expand&Shrink&mAP\\
\hline\hline
Naive& & & & 26.6  \\
Static&\checkmark& & & 34.4  \\
Static$+$expand&\checkmark&\checkmark& & 35.0 \\
Static$+$shrink&\checkmark&&\checkmark& 34.8 \\
Dynamic&\checkmark&\checkmark&\checkmark&\bfseries{35.5} \\
\hline
\end{tabular}
\end{center}
\caption{Detection performance (mAP in \%) of models with different operations.
\textit{``Pseudo-label''} means using submodels to generate missing annotations.
\textit{``Expand''} and \textit{``Shrink''} mean updating on the annotation sets.
}
\label{dynamic-supervisor}
\end{table}

\begin{table}[t]
\begin{center}
\begin{tabular}{c | c c |c}
\hline
 Operation&$\Delta$Recall(\%)&$\Delta$Precision(\%)&mAP\\
\hline\hline
Pseudo-label&0&0&34.4\\
\hline\hline
Expand&\textcolor[RGB]{2,167,71}{+27.6}&\textcolor{red}{-5.2}&35.0\\
Expand-shrink&\textcolor[RGB]{2,167,71}{+16.8}&\textcolor[RGB]{2,167,71}{+25.9}&\textbf{35.5}\\
\hline\hline
Shrink&\textcolor{red}{-6.0}&\textcolor[RGB]{2,167,71}{+36.6}&34.8\\
Shrink-expand&\textcolor[RGB]{2,167,71}{+23.1}&\textcolor[RGB]{2,167,71}{+9.8}&35.2\\
\hline
\end{tabular}
\end{center}
\caption{The relative variation of recall and precision when new submodels expand or shrink the original pseudo annotations set.
}
\label{p-r-change}
\end{table}

\textbf{Dynamic Supervisor \textit{vs.} Static Supervisor:}
We report the detection performance of models with different operations in Table~\ref{dynamic-supervisor}.
The model trained using the naive combination of two datasets, without any other strategies, performs the worst.
When the static supervisor framework is applied, the corresponding model enhances the performance by 7.8\% in mAP, showing the effect of pseudo-labeling on the cross-dataset object detection task.
Furthermore, a dynamic supervisor framework is proposed to update the initial annotation sets multiple times using two types of submodels.
From the last three models shown in Table~\ref{dynamic-supervisor}, the promotions resulting from the expanding operation, the shrinking operation, or both are clear.
In Table~\ref{p-r-change}, we record the variation in recall and precision when new submodels apply expanding, shrinking, or a combination of both to the annotation set.
A single expanding operation on the annotation sets increases the recall but decreases the precision, and the shrinking operation is simply the opposite.
Nevertheless, when we combine the advantages of these two operations and apply them sequentially, both the recall and precision can be improved.

\begin{table}[t]
\begin{center}
\begin{tabular}{c|c c|c}
\hline
Type                             & Expand & Shrink & mAP \\
\hline\hline
\multirow{2}{*}{Self-annotated}  &\checkmark&          &34.8 \\
                                 &\checkmark&\checkmark&34.6 \\ \hline\hline
\multirow{2}{*}{Cross-annotated} &\checkmark&          &35.0 \\
                                 &\checkmark&\checkmark&\textbf{35.5} \\ \hline
\end{tabular}
\end{center}
\caption{Detection performance (mAP in \%) of different types of dynamic supervisor.
\textit{``Expand''} and \textit{``Shrink''} mean the update operation on pseudo annotations.
}
\label{cross-self}
\end{table}

\begin{figure*}
\begin{center}
\includegraphics[width=1.0\linewidth]{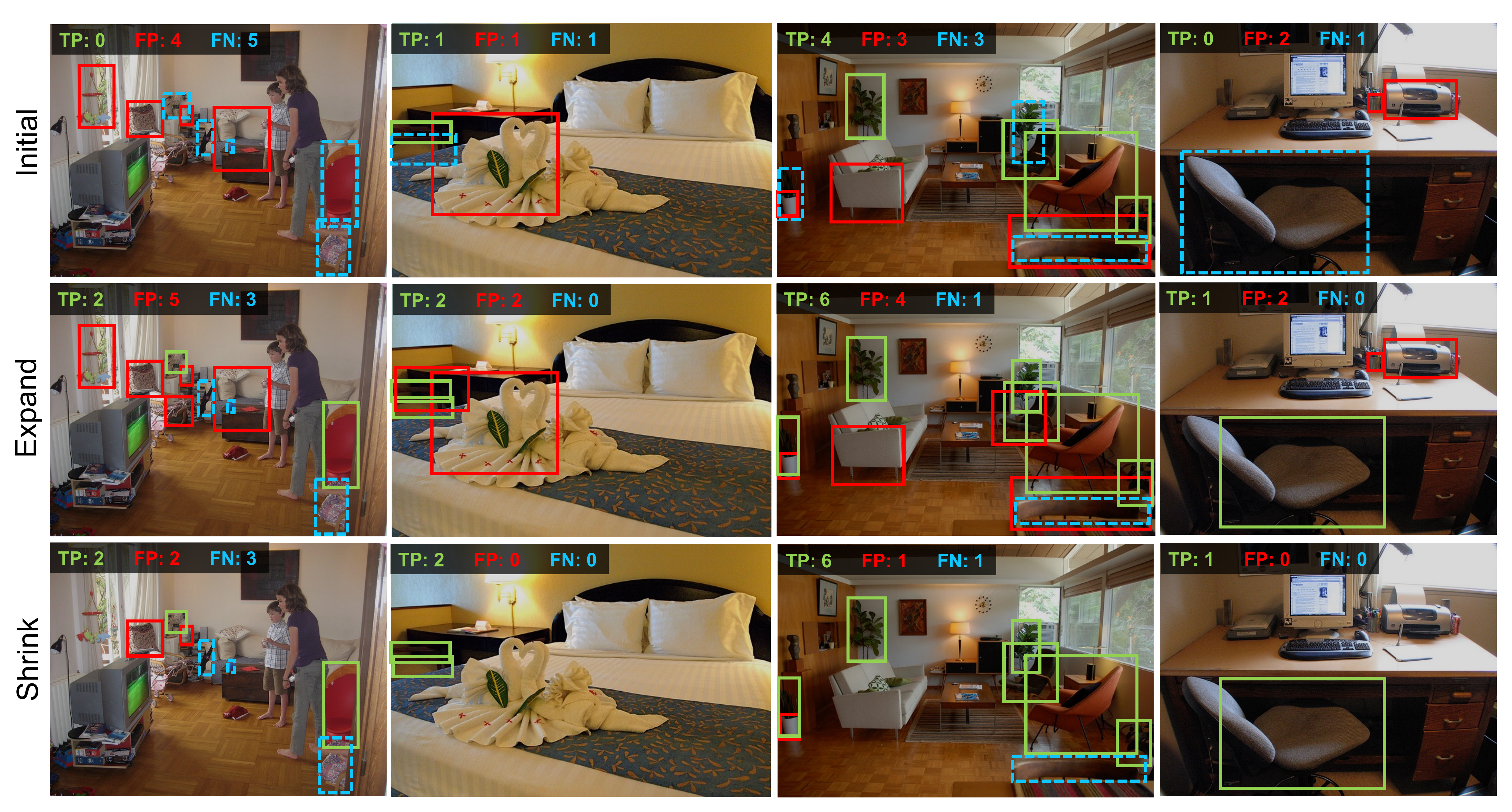}
\end{center}
   \caption{Visualizations of pseudo-labeling results in different steps of the dynamic supervisor framework.
   These images are sampled from the \textit{miniCOCO-beta} and their original ground truth boxes are not labeled here.
   Green rectangle indicates true annotations, red rectangle indicates false annotations, and blue dotted rectangle indicates missing annotations.
   The numbers of true annotations (TP), false annotations (FP), and missing annotations (FN) are listed inside every image.
   }
\label{fig:qualitative}
\end{figure*}

\textbf{Cross-annotated \textit{vs.} Self-annotated:}
In Section~\ref{section4-1}, we introduce two types of the dynamic supervisor framework, which is a ``self-annotated mechanism'' for the first one and a ``cross-annotated mechanism'' for the second one.
In the self-annotated mechanism, two datasets are merged, and the detection model is trained using this merged dataset to expand or shrink the initial annotation sets after they are generated. 
In the cross-annotated mechanism, there are two models trained using their respective datasets, which are already augmented with annotation sets, after which they are utilized to expand or shrink the annotation set of each other.
The detection performance of these two types of dynamic supervisors is shown in Table~\ref{cross-self}.
When the expanding operation is applied to the annotation set, both mechanisms result in performance improvements.
However, when the shrinking operation is applied subsequently, the detection performance of the self-annotated mechanism tends to degrade (mAP decreases from 34.8\% to 34.6\%).
We suggest that the model trained using a noisy dataset is prone to overfitting to incorrect annotations.
Consequently, in the self-annotated mechanism, the knowledge learned from such a noisy dataset is that it is difficult to eliminate the noise of this dataset continuously.
In contrast, the cross-annotated mechanism avoids this dilemma and can progressively improve performance (mAP increases from 34.4\% to 35.5\% step by step).

\textbf{Operation Sequence:}
There are two operations in the proposed dynamic supervisor framework for improving the recall or precision of the annotation set.
Here, we attempt to update the annotation set through a different sequence of operations to explore the relationship between quality variation and performance improvement.
The results of the experiments on the two types of operation sequences are shown in Table~\ref{p-r-change}.
First, and most importantly, the two types of operation sequences can improve the quality of the annotation set and achieve similar detection performance (35.5\% \textit{vs.} 35.2\%).
However, the different operation sequences would result in different improvements in recall and precision.
When we first expand the annotation set and then shrink it, we obtain more improvement in precision and less improvement in recall.
This is because the later shrinking operation is prone to missing TP. 
Therefore, it suppresses recall.
Conversely, the improvement in the recall will be higher.
The results show the flexibility of the dynamic supervisor framework, and the goal is to improve the quality of the annotation set comprehensively.

\subsection{Qualitative Visualizations:}
Finally, we visualize the pseudo-labeling results in different steps of our dynamic supervisor framework, as shown in Figure~\ref{fig:qualitative}.
These images are sampled from \textit{miniCOCO-Beta}, and their original ground-truth boxes are not labeled here.
As described in previous sections, the pseudo-labeling results of a single model are biased, and it is difficult for a single model to achieve high recall and high precision (the first row of Figure~\ref{fig:qualitative}).
Therefore, we propose updating the annotations set multiple times in a dynamic framework.
The expanding operation is effective for increasing the number of TP. However, it also facilitates the introduction of new FP (the second row of Figure~\ref{fig:qualitative}).
Consequently, the shrinking operation is applied to discard them and obtain a cleaner annotation set (the third row of Figure~\ref{fig:qualitative}).
However, compared with the ground truth boxes, objects, such as the cup (on the table) and handbag (hanging on the sofa), which are small or hard to distinguish, still need to be found.

\section{Conclusion}
In this study, to address the problem of cross-dataset object detection, we reveal the implicit connections between hard- and soft-label training and the methods for pseudo-annotation ensembling.
We show that hard-label training and soft-label training are conducive to improving the recall and precision of the detection results, respectively.
Based on this, we propose a dynamic supervisor framework, which polishes the annotations dynamically and selects predictions adaptively based on category.
The proposed dynamic supervisor framework updates the annotation set multiple times and improves its quality progressively.
Experiments conducted on different combinations of several datasets demonstrate the effectiveness of the proposed dynamic supervisor framework.

\section{Acknowledgements}
This work was supported by Alibaba Group through Alibaba Research Intern Program.
And this work was supported by the Fundamental Research Funds for the Central Universities, and the National Natural Science Foundation of China under Grant 31627802.



\bibliographystyle{elsarticle-num}
\bibliography{mybibfile}


%\begin{thebibliography}{00}


%\bibitem{}

%\end{thebibliography}
\end{document}